\pdfoutput=1

\documentclass[11pt]{article}

\usepackage[]{emnlp2021}

\usepackage{times}
\usepackage{latexsym}
\usepackage{graphicx}
\usepackage{makecell}
\usepackage{multirow}
\usepackage{amsmath}
\usepackage{amsfonts}

\usepackage[T1]{fontenc}

\usepackage[utf8]{inputenc}

\usepackage{microtype}

\title{Graph Based Network with Contextualized Representations of Turns in Dialogue}

\author{Bongseok Lee \and Yong Suk Choi\Thanks{ Corresponding author}\\ 
  Department of Computer Science \\
  Hanyang University, Seoul, Korea \\
  \texttt{\{tjr4090, cys\}@hanyang.ac.kr} \\}

\begin{document}
\maketitle
\begin{abstract}
Dialogue-based relation extraction (RE) aims to extract relation(s) between two arguments that appear in a dialogue. Because dialogues have the characteristics of high personal pronoun occurrences and low information density, and since most relational facts in dialogues are not supported by any single sentence, dialogue-based relation extraction requires a comprehensive understanding of dialogue. In this paper, we propose the \textbf{TU}rn \textbf{CO}ntext awa\textbf{RE} \textbf{G}raph \textbf{C}onvolutional \textbf{N}etwork (TUCORE-GCN) modeled by paying attention to the way people understand dialogues. In addition, we propose a novel approach which treats the task of emotion recognition in conversations (ERC) as a dialogue-based RE. Experiments on a dialogue-based RE dataset and three ERC datasets demonstrate that our model is very effective in various dialogue-based natural language understanding tasks. In these experiments, TUCORE-GCN outperforms the state-of-the-art models on most of the benchmark datasets. Our code is available at \url{https://github.com/BlackNoodle/TUCORE-GCN}.
\end{abstract}

\section{Introduction}
The task of relation extraction (RE) aims to identify semantic relations between arguments from a text, such as a sentence, a document, or even a dialogue. However, since a large number of relational facts are expressed in multiple sentences, sentence-level RE suffers from inevitable restrictions in practice \citep{yao-etal-2019-docred}. Therefore, cross-sentence RE, which aims to identify relations between two arguments that are not mentioned in the same sentence or relations that cannot be supported by any single sentence, is an essential step in building knowledge bases from large-scale corpora automatically \citep{Ji10overviewof, swampillai-stevenson-2010-inter, Surdeanu2013OverviewOT}. In this respect, because dialogues readily exhibit cross-sentence relations \citep{yu-etal-2020-dialogue}, extracting relations from the dialogue is necessary.

To support the prediction of relation(s) between two arguments that appear within a dialogue, \citet{yu-etal-2020-dialogue} recently proposed DialogRE, which is a human-annotated dialogue-based RE dataset.
Table~\ref{dialogRE-table} shows an example of DialogRE. 
In conversational texts such as DialogRE, because of its higher person pronoun frequency \citep{Biber_Variation88} and lower information density \citep{wang-liu-2011-pilot} compared to formal written texts, most relational triples require reasoning over multiple sentences in a dialogue. 
65.9\% of relational triples in DialogRE involve arguments that never appear in the same turn. Therefore, multi-turn information plays an important role in dialogue-based RE.

\begin{table}
\centering
{\small
\begin{tabular}{cl}
\Xhline{3\arrayrulewidth}
\textbf{S1:} & Hey Pheebs. \\
\textbf{S2:} & Hey! \\
\textbf{S1:} & Any sign of your brother? \\
\textbf{S2:} & No, but he’s always late. \\
\textbf{S1:} & I thought you only met him once? \\
\textbf{S2:} & Yeah, I did. I think it sounds y’know big sistery,\\
 & y’know, ‘Frank’s always late.’ \\
\textbf{S1:} & Well relax, he’ll be here. \\
\Xhline{2\arrayrulewidth}
\multicolumn{2}{l}{\textbf{Subject:} Frank} \\
\multicolumn{2}{l}{\textbf{Object:} S2} \\
\multicolumn{2}{l}{\textbf{relation:} per:siblings} \\
\hline
\multicolumn{2}{l}{\textbf{Subject:} S2} \\
\multicolumn{2}{l}{\textbf{Object:} Frank} \\
\multicolumn{2}{l}{\textbf{relation:} per:siblings} \\
\hline
\multicolumn{2}{l}{\textbf{Subject:} S2} \\
\multicolumn{2}{l}{\textbf{Object:} Pheeb} \\
\multicolumn{2}{l}{\textbf{relation:}per:alternate\_names} \\
\Xhline{3\arrayrulewidth}
\end{tabular}
}
\caption{\label{dialogRE-table} An example dialogue and its desired relations in DialogRE \citep{yu-etal-2020-dialogue}. S1, S2: anonymized speaker of each utterance. }
\end{table}

There are several major challenges in effective relation extraction from dialogue, inspired by the way how people understand dialogue in practice. First, the dialogue has speakers, and who speaks each utterance matters. The reason for it is because the subject and object of relational triples depend on who is speaking which utterance. For example, if S3 answered \textit{``Hey!''} after \textit{``Hey Pheebs.''}, the relational triple $($S2, per:alternate\_names, Pheebs$)$ will be revised to $($S3, per:alternate\_names, Pheebs$)$, in the case of Table~\ref{dialogRE-table}. Second, when understanding the meaning of each turn in a dialogue, it is important to know the meaning of the surrounding turns. For example, if we look at \textit{``No, but he is always late.''} in Table~\ref{dialogRE-table}, we don't know who's always late. However, if we look at the previous turn, we can see that S2's brother is always late. Third, the dialogue consists of several turns. Those turns are sequential, and the arguments may appear in different turns. Consequently, it is important to grasp the multi-turn information in order to capture the relations between the two arguments. This could be done using the sequential characteristics of dialogues. Therefore, we aim to tackle these challenges to better extract relations from dialogues.

In this paper, we propose the \textbf{TU}rn \textbf{CO}ntext awa\textbf{RE} \textbf{G}raph \textbf{C}onvolutional \textbf{N}etwork (TUCORE-GCN) for dialogue-based RE. It is designed to tackle the aforementioned challenges. TUCORE-GCN encodes the input sequence to reflect speaker information in dialogue by applying BERT$_s$ \citep{yu-etal-2020-dialogue} and speaker embedding of SA-BERT \citep{gu2020speaker}. Then, to better extract the representations of each turn from the encoded input sequence, Masked Multi-Head Self-Attention \citep{NIPS2017_3f5ee243} is applied using a surrounding turn mask. Next, TUCORE-GCN constructs a heterogeneous dialogue graph to capture the relational information between arguments in the dialogue. It consists of four types of nodes, namely dialogue node, turn node, subject node, object node, and three different types of edges, i.e., speaker edge, dialogue edge, and argument edge. Then, the sequential characteristics of the turn nodes should be considered. To obtain a surrounding turn-aware representation for each node, we apply bidirectional LSTM (BiLSTM) \citep{650093} to the turn nodes and a Graph Convolutional Network \citep{Kipf:2016tc} to the heterogeneous dialogue graph. Finally, we classify the relations between arguments with the obtained features.

The task of emotion recognition in conversations (ERC) aims to identify the emotion of utterances in dialogue. ERC is a challenging task that has recently gained popularity due to its potential applications \citep{poria-etal-2019-meld}. It can be used to analyze user behaviors \citep{LEE2016360} and detect fake news \citep{guo2019dean}. Table~\ref{EmoryNLP-table} shows an example from EmoryNLP \citep{DBLP:conf/aaai/ZahiriC18}, a dataset widely used in the ERC task. We propose a novel approach to treat the ERC task as a dialogue-based RE. 
If we define the emotion relation of each utterance when the subject says the object with a particular emotion (e.g., joyful, neutral, scared), the emotion of each utterance in the dialogue can be seen as a triple (speaker of utterance, emotion, utterance) as shown in Table~\ref{erd2dialogRE-table}. To the best of our knowledge, this approach was not introduced in previous studies.

\begin{table}
\centering
{\small
\begin{tabular}{clc}
\hline \textbf{Speaker} & \textbf{Utterance} & \textbf{Emotion} \\ \hline
Monica & He is so cute. So, where did & Joyful \\
& you guys grow up? & \\
Angela & Brooklyn Heights. & Neutral \\
Bob & Cleveland. & Neutral \\
Monica & How, how did that happen? & Neutral \\
Joey & Oh my god. & Scared \\
Monica & What? & Neutral \\
Joey & I suddenly had the feeling that & Scared \\
& I was falling. But I'm not. & \\
\hline
\end{tabular}
}
\caption{\label{EmoryNLP-table} An example conversation with annotated labels in EmoryNLP\citep{DBLP:conf/aaai/ZahiriC18}.}
\end{table}

\begin{table}
\centering
{\small
\begin{tabular}{cl}
\Xhline{3\arrayrulewidth}
\textbf{S1:} & He is so cute. So, where did you guys grow up? \\
\textbf{S2:} & Brooklyn Heights. \\
\textbf{S3:} & Cleveland. \\
\textbf{S1:} & How, how did that happen? \\
\textbf{S4:} & Oh my god. \\
\textbf{S1:} & What? \\
\textbf{S4:} & I suddenly had the feeling that I was falling. But \\
& I'm not. \\
\Xhline{2\arrayrulewidth}
\multicolumn{2}{l}{\textbf{Subject:} S1} \\
\multicolumn{2}{l}{\textbf{Object:} He is so cute. So, where did you guys grow up?} \\
\multicolumn{2}{l}{\textbf{relation:} Joyful} \\
\hline
\multicolumn{2}{l}{\textbf{Subject:} S2} \\
\multicolumn{2}{l}{\textbf{Object:} Brooklyn Heights.} \\
\multicolumn{2}{l}{\textbf{relation:} Neutral} \\
\hline
\multicolumn{2}{l}{\textbf{Subject:} S3} \\
\multicolumn{2}{l}{\textbf{Object:} Cleveland.} \\
\multicolumn{2}{l}{\textbf{relation:} Neutral} \\
\Xhline{3\arrayrulewidth}
\end{tabular}
}
\caption{\label{erd2dialogRE-table} An example of converting the example in Table~\ref{EmoryNLP-table} to DialogRE format to treat the ERC task as a dialogue-based RE. S1, S2, S3, S4: anonymized speaker of each utterance.}
\end{table}

In summary, our main contributions are as follows:
\begin{itemize}
\item We propose a novel method, TUrn COntext awaRE Graph Convolutional Network (TUCORE-GCN), to better cope with a dialogue-based RE task.
\item We introduce a surrounding turn mask to better capture the representation of the turns.
\item We introduce a heterogeneous dialogue graph to model the interaction among elements (e.g., speakers, turns, arguments) across the dialogue and propose a GCN mechanism combined with BiLSTM.
\item We propose a novel approach to treat the ERC task as a dialogue-based RE.
\end{itemize}

\section{Related Work}
\subsection{Dialogue-Based Relation Extraction} 
Relation extraction has been studied extensively over the past few years and many approaches have achieved remarkable success. Most previous approaches focused on sentence-level RE \citep{zeng-etal-2014-relation, wang-etal-2016-relation, zhang-etal-2017-position, zhu-etal-2019-graph}, but recently cross-sentence RE has been studied more because a large number of relational facts are expressed in multiple sentences in practice.

Recent work begins to explore cross-sentence relation extraction on documents that are formal genres, such as professionally written and edited news reports or well-edited websites. In document-level RE, various approaches including transformer-based methods \citep{DBLP:conf/pakdd/TangC0CFWY20, Ye2020CoreferentialRL, DBLP:journals/corr/abs-1909-11898} and graph-based methods \citep{christopoulou-etal-2019-connecting, nan-etal-2020-reasoning, zeng-etal-2020-double} have been proposed. Among these, graph-based methods are widely adopted in document-level RE due to their effectiveness and strength in representing complicated syntactic and semantic relations among structured language data. Unlike previous work, we focused on extracting relations from dialogues, which are texts with high pronoun frequencies and low information density.

\citep{yu-etal-2020-dialogue, DBLP:conf/aaai/XueSZC21} were among the early works on dialogue-based RE. \citet{yu-etal-2020-dialogue} introduced several dialogue-based RE approaches with the DialogRE dataset. Among the various approaches, BERT$_s$, a model that uses BERT \citep{devlin-etal-2019-bert}, shows good performance. BERT$_s$ is a model that slightly modified the original input sequence of BERT in consideration of speaker information. However, it has a limitation in that it cannot predict asymmetric inverse relations well. 
Our model basically follows the input sequence of BERT$_s$, but we designed it to overcome this limitation to some extent. More detailed explanation is in Sec~\ref{ssec:Analysis}. \citet{DBLP:conf/aaai/XueSZC21} proposed a graph-based approach, GDPNet, that constructs a latent multi-view graph to capture various possible relationships among tokens and refines this graph to select important words for relation prediction. In this approach, the refined graph and the BERT-based sequence representations are concatenated for relation extraction. 
The graph of GDPNet is a multi-view directed graph aiming to model all possible relationships between tokens. Unlike GDPNet, we combine tokens into meaningful units to form nodes and connect the nodes with speaker edges, dialogue edges, and argument edges to model what each edge means. In addition, GPDNet focuses on refining this multi-view graph to capture important words from long texts for RE, but we extract the relations using the features of the nodes in the graph.

\subsection{Emotion Recognition in Conversation}
Emotion recognition in conversation has emerged as an important problem in recent years and many successful approaches have been proposed. In ERC, numerous approaches including recurrence-based methods \citep{MajumderPHMGC19, ghosal-etal-2020-cosmic} and graph-based methods \citep{ghosal-etal-2019-dialoguegcn, ishiwatari-etal-2020-relation} have been proposed. For instance, DialogueRNN \citep{MajumderPHMGC19} uses an attention mechanism to grasp the relevant utterance from the whole conversation and models the party state, global state, and emotional dynamics with several RNNs. COSMIC \citep{ghosal-etal-2020-cosmic} adopts a network structure, which is similar to DialogueRNN but adds external common sense knowledge to improve performance. DialogueGCN \citep{ghosal-etal-2019-dialoguegcn} treats each dialogue as a graph where each node represents utterance and is connected to the surrounding utterances. RGAT \citep{ishiwatari-etal-2020-relation} is based on DialogueGCN. It adds relational positional encodings that can capture speaker dependency, along with sequential information. Many studies with remarkable success have been proposed, but none can be used in ERC as well as other dialogue-based tasks like our approaches.

\begin{figure*}
\begin{center}
\includegraphics[width=\linewidth]{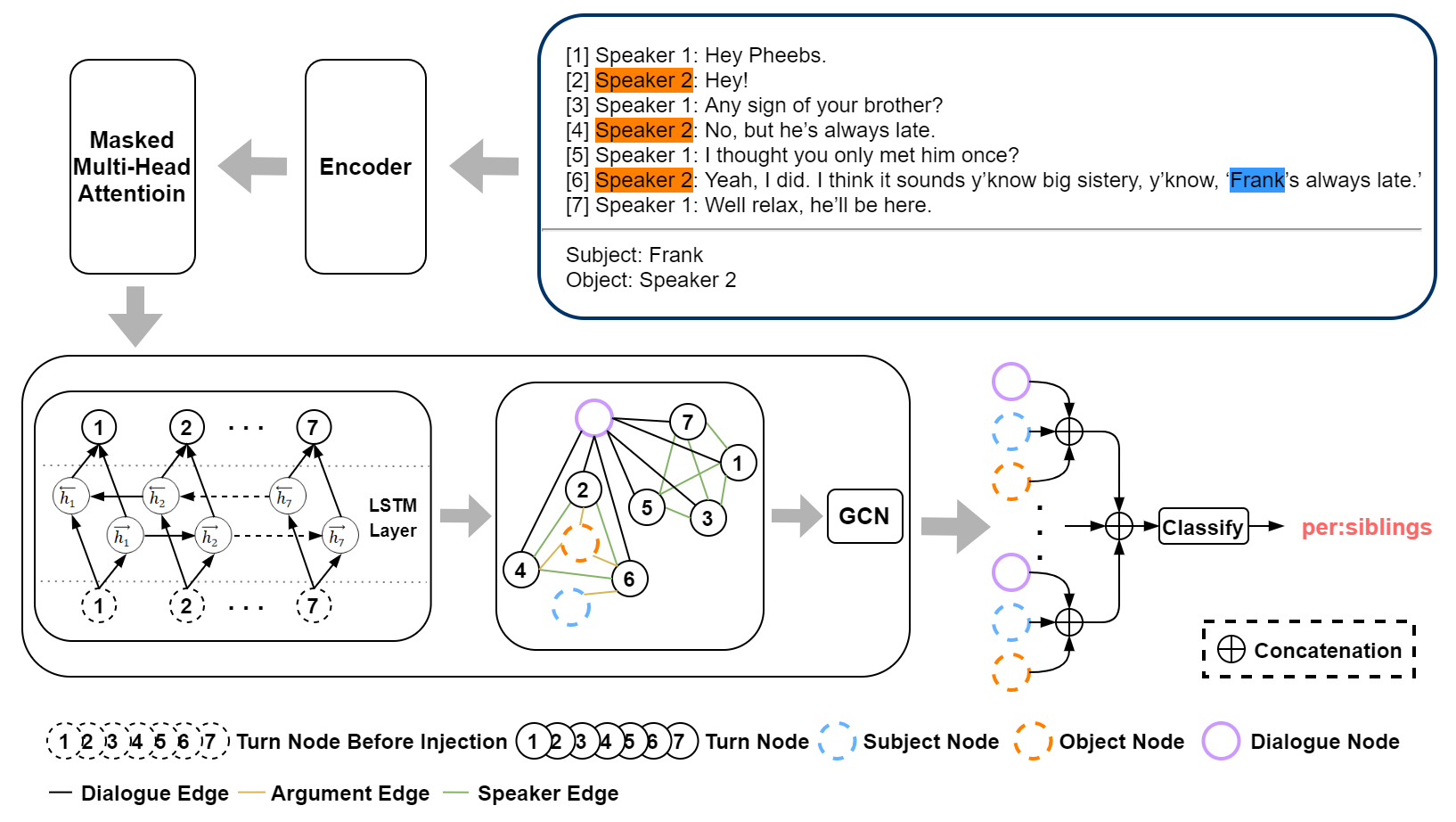}
\end{center}
\caption{The overall architecture of TUCORE-GCN. First, A contextualized representation of each token is obtained by feeding the input dialogue to the context encoder. Next, Masked Multi-Head Attention using surrounding turn mask is applied to obtain representations that enhance the meaning of each turn. Then, TUCORE-GCN constructs a dialogue graph and applies GCN mechanism combined with BiLSTM. Finally, the classification module predicts relations using information from the previous module.}
\label{fig:architecture}
\end{figure*}

\section{Model}
TUCORE-GCN mainly consist of four modules: encoding module (Sec~\ref{ssec:Encoding Module}), turn attention module (Sec~\ref{ssec:Turn Attention Module}), dialogue graph with sequential nodes module (Sec~\ref{ssec:Dialog Graph With Sequential Nodes Module}), and classification module (Sec~\ref{ssec:Classification Module}), as shown in Figure~\ref{fig:architecture}.

\subsection{Encoding Module}
\label{ssec:Encoding Module}
We follow BERT$_s$ \citep{yu-etal-2020-dialogue} as the input sequence of the encoding module. Given a dialogue $d = s_1:t_1,s_2:t_2,...,s_M:t_M$ and its associated argument pair $(a_1, a_2)$, where $s_i$ and $t_i$ denote the speaker ID and text of the $i^{th}$ turn, respectively, and $M$ is the total number of turns, BERT$_s$ constructs $\hat{d} = \hat{s}_1:t_1,\hat{s}_2:t_2,...,\hat{s}_M:t_M$, where $\hat{s}_i$ is:
\begin{equation}
    \hat{s}_i =\begin{cases} [S_1] & \text{if } s_i = a_1 \\ [S_2] & \text{if } s_i = a_2 \\ \hfil s_i & \text{otherwise}\end{cases} 
\end{equation}
where $[S_1]$ and $[S_2]$ are special tokens. In addition, it defines $\hat{a}_k (k \in \{1,2\})$ to be \texttt{$[S_k]$} if $\exists i (s_i=a_k)$, and $a_k$ otherwise. Then, we concatenate $\hat{d}$ and $(\hat{a}_1, \hat{a}_2)$ with a classification token \texttt{[CLS]} and a separator token \texttt{[SEP]} in BERT\citep{devlin-etal-2019-bert} as the input sequence \texttt{[CLS]$\hat{d}$[SEP]$\hat{a}_1$[SEP]$\hat{a}_2$[SEP]}.

To model the speaker change information, following SA-BERT\citep{gu2020speaker}, we add additional speaker embeddings to the token representations. $E_s(\hat{s}_i)$ is added to each token representation of $\hat{s}_i:t_i$, $E_s(\hat{a}_k) (k \in \{1,2\})$ is added to each token representation of $\hat{a}_k$ if $\hat{a}_k = [S_k]$, and $E_s(\sharp)$ is added to all token representations without speaker embedding added, where $E_s(\cdot)$ denotes the speaker embedding layer. $E_s(\sharp)$ is an embedding output for token representations without speaker information. A visual architecture of our input representation is illustrated in Appendix.

Then, token representations containing speaker change information are fed into an encoder to extract the speaker-sensitive token representations. The encoder can be BERT or BERT variants \citep{liu2019roberta, NEURIPS2019_c04c19c2, Lan2020ALBERT:}.


\subsection{Turn Attention Module}
\label{ssec:Turn Attention Module}
To obtain the turn context-sensitive representation for each turn, we apply Masked Multi-Head Self-Attention \citep{NIPS2017_3f5ee243} to the output of the encoder using the surrounding turn mask. The range of this surrounding turn is called the window, and the number of turns from the front and rear are viewed as the surrounding turn which is called the surround turn window size. The surround turn window size $c$ is a hyper-parameter.

Let $X = [x_1,x_2,x_3,...,x_N]$ be an output of the encoding module, where $x_j$ is the $j^{th}$ token representation in the output and N is the number of tokens. For token representations corresponding to $\hat{s}_i:t_i$ range from $x_{b_i}$ to $x_{e_i}$, $D_i = [x_{b_i},x_{b_i+1},...,x_{e_i}]$ denotes representations of the $i^{th}$ turn, $D = [x_{b_1},x_{b_1+1},...,x_{e_M}]$ denotes representations of a dialogue $\hat{d}$, and $F(x_m)$ denotes the turn number in which $x_m$ is included (e.g., $F(x_m) = 2$ if $x_m \in D_2$). We implement the surrounding turn mask as follows:
\begin{equation}
    M^{sur}_{mn} = \begin{cases} \hfil 1 & \text{if } x_m \notin D, m = n \\
                                 \hfil -\infty & \text{if } x_m \notin D, m \neq n \\ 
                                 \hfil 1 & \text{if } x_m \in D, x_n \in R(x_m) \\
                                 \hfil -\infty & \text{otherwise}\end{cases}
\end{equation}
where $R(x_m)$ denotes $\bigcup\limits_{z=F(x_m)-c}^{F(x_m)+c} D_{z}$. A visual architecture of an example regarding the surrounding turn mask is illustrated in Appendix.

Then, we reinforce the representation of each turn from representations of surrounding turns.

\subsection{Dialogue Graph with Sequential Nodes Module}
\label{ssec:Dialog Graph With Sequential Nodes Module}
To model the dialogue-level information, interactions between turns and arguments, and interactions between turns, a heterogeneous dialogue graph is constructed. 

We form four distinct types of nodes in the graph: dialogue node, turn node, subject node, and object node. The dialogue node is a node with the purpose of containing overall dialogue information. Turn nodes represent information about each turn in the dialogue and are created as many as the total number of turns in the dialogue. The subject node and object node represent the information of each argument. In our work, the initial representation of the dialogue node uses a feature corresponding to \texttt{[CLS]} in the output of the turn attention module. The initial representation of the $i^{th}$ turn node, subject node, and object node use the average of the token representations corresponding to \texttt{$\hat{s}_i:t_i$}, \texttt{$\hat{a}_1$}, and \texttt{$\hat{a}_2$} in the output of the turn attention module, respectively. 

There are three different types of edge:
\begin{itemize}
\item \textbf{dialogue edge: } All turn nodes are connected to the dialogue node with the dialogue edge so that the dialogue node learns while being aware of turn-level information.
\item \textbf{argument edge: } To model the interaction between turns and arguments, the $i^{th}$ turn node and argument nodes (i.e., subject node and object node) are connected with the argument edge if the argument is mentioned in \texttt{$\hat{s}_i:t_i$}.
\item \textbf{speaker edge:} To model the interaction among different turns of the same speaker, turn nodes uttered by the same speaker are fully connected with speaker edges. 
\end{itemize}

Next, we apply a Graph Convolutional Network (GCN) \citep{Kipf:2016tc} to aggregate each node feature from the features of the neighbors. At this time, in order to inject sequential information to the turn nodes, GCN is applied after the turn nodes pass through the bidirectional LSTM \citep{650093} layers.
Given node $u$ at the $l^{th}$ GCN layer, $h^{(l)}_u$ and $\hat{h}^{(l)}_u$ denote the representation of the node before injecting sequential information and the representation of the node after injecting sequential information, respectively.
$\hat{h}^{(l)}_u$ can be defined as:
\begin{equation}
    \dot{h}^{(l)}_{T_i} = [\overrightarrow{LSTM^{(l)}}(h^{(l)}_{T_i});\overleftarrow{LSTM^{(l)}}(h^{(l)}_{T_i})]
\end{equation}
\begin{equation}
    \hat{h}^{(l)}_u = \begin{cases} W^{(l)}_{\alpha}\dot{h}^{(l)}_{T_i} + b^{(l)}_{\alpha} & \text{if type of }u \text{ is turn node}\\ \\ \hfil h^{(l)}_u & \text{otherwise} \end{cases}
\end{equation}
where $T_i$ represents an $i^{th}$ turn node and $\dot{h}^{(l)}_{T_i}$ represents turn node feature injected sequential information in the dialogue by concatenating the hidden states of two directions. $W^{(l)}_{\alpha} \in \mathbb{R}^{d \times 2d}$, $b^{(l)}_{\alpha} \in \mathbb{R}^{d}$, and $d$ is the dimension. Then, the graph convolution operation can be defined as:
\begin{equation}
    h^{(l+1)}_u = ReLU\left(\sum_{k \in \kappa} \sum_{v \in N_k(u)} W^{(l)}_k\hat{h}^{(l)}_v+b^{(l)}_k\right)
\end{equation}
where $\kappa$ are different types of edges, $N_k(u)$ denotes neighbors for node $u$ connected in the $k^{th}$ type edge, $W^{(l)}_{k} \in \mathbb{R}^{d \times d}$, and $b^{(l)}_{k} \in \mathbb{R}^{d}$.

\subsection{Classification Module}
\label{ssec:Classification Module}

We concatenate the dialogue node, subject node, and object node to classify the relation between arguments.
Furthermore, to cover features of all different abstract levels from each layer of the GCN, we concatenate the hidden states of each GCN layer as follows:
\begin{equation}
    C = [h^{(0)}_d;h^{(0)}_s;h^{(0)}_o;...;h^{(G)}_d;h^{(G)}_s;h^{(G)}_o;]
\end{equation}
where $G$ is the number of GCN layers and $d$, $s$, and $o$ denote the dialogue node, subject node, and object node, respectively.
For each relation type $r$, we introduce a vector $W_r \in \mathbb{R}^{3(G+1)d}$ and obtain the probability $P_r$ of the existence of $r$ between arguments by $P_r = sigmoid(CW^T_r)$. We use cross-entropy loss as the classification loss to train our model in an end-to-end way.

\section{Experiments}
In this section, we report our experimental results on two tasks, dialogue-based RE and ERC. We experiment with two versions of TUCORE-GCN, TUCORE-GCN$_{BERT}$ and TUCORE-GCN$_{RoBERTa}$, respectively based on the uncased base model of BERT \citep{devlin-etal-2019-bert} and the large model of RoBERTa \citep{liu2019roberta}. TUCORE-GCN is trained using Adam \citep{DBLP:journals/corr/KingmaB14} as an optimizer with weight decay 0.01. We run each experiment five times and report the average score along with the standard deviation ($\sigma$) for each metric. The full details of our training settings are provided in the Appendix.

\subsection{Dialogue Based Relation Extraction}
\subsubsection{Dataset}
We evaluate our model on DialogRE \citep{yu-etal-2020-dialogue}, an updated English version with a few annotation errors fixed \footnote{https://dataset.org/dialogre}. DialogRE has 36 relation types, 1,788 dialogues, and 8,119 triples, not including no-relation argument pairs, in total. We follow the standard split of the dataset.
\subsubsection{Metrics}
For DialogRE, We calculate both the $F1$ and $F1_c$ \citep{yu-etal-2020-dialogue} scores as the evaluation metrics. $F1_c$ is an evaluation metric to supplement the standard $F1$. $F1_c$ is computed by taking in the part of dialogue as input, instead of only considering the entire dialogue.
\subsubsection{Baselines and State-of-the-Art}
\label{ssec:RE-baseline}
For a comprehensive performance evaluation, we compared our model with the models using the following baseline and state-of-the-art methods: 

\textbf{BERT \citep{devlin-etal-2019-bert}:} The BERT baseline for dialog-based RE, initialized with pre-trained parameters of BERT-base. It is classified using a final hidden vector corresponding to the [CLS] token. 

\textbf{BERT$_s$ \citep{yu-etal-2020-dialogue}:} A modification to the input sequence of the above BERT baseline. This modification prevents a model from overfitting to the training data and helps a model locate the start positions of relevant turns. 

\textbf{GDPNet \citep{DBLP:conf/aaai/XueSZC21}:} A state-of-the-art model for the DialogRE. GDPNet finds indicative words from long sequences by constructing a latent multi-view graph and refining the graph. It uses the same input format of BERT$_s$ and pre-trained parameters of BERT-base. 

\textbf{RoBERTa$_s$:} A model that uses the pre-trained parameters of RoBERTa-large \citep{liu2019roberta} instead of pre-trained parameters of the BERT$_s$ above.

\begin{table*}
\centering
{\small
\begin{tabular}{c|cc|cc}
\Xhline{3\arrayrulewidth}
\multirow{2}{*}{\textbf{Method}} & \multicolumn{2}{c|}{\textbf{Dev}} & \multicolumn{2}{c}{\textbf{Test}} \\
& F1 ($\sigma$) & F1$_c$ ($\sigma$) & F1 ($\sigma$) & F1$_c$ ($\sigma$) \\
\hline
BERT & 59.4 (0.7) & 54.7 (0.8) & 57.9 (1.0) & 53.1 (0.7) \\
BERT$_s$ & 62.2 (1.3) & 57.0 (1.0) & 59.5 (2.1) & 54.2 (1.4) \\
GDPNet & 61.8 (1.4)* & 58.5 (1.4)* & 60.2 (1.0)* & 57.3 (1.2)* \\
RoBERTa$_s$ & 72.6 (1.7) & 65.1 (1.7) & 71.3 (1.6) & 63.7 (1.2) \\
\hline
TUCORE-GCN$_{BERT}$ & 66.8 (0.7) & 61.0 (0.5) & 65.5 (0.4) & 60.2 (0.6) \\
TUCORE-GCN$_{RoBERTa}$ & \textbf{74.3} (0.6) & \textbf{67.0} (0.6) & \textbf{73.1} (0.4) & \textbf{65.9} (0.6) \\
\Xhline{3\arrayrulewidth}
\end{tabular}
}
\caption{\label{dialogRE-result} Performance on DialogRE. The scores marked by “*” are based on our re-implementation, because of the data differences.}
\end{table*}

\subsubsection{Results}
We show the performance of TUCORE-GCN on the DialogRE dataset in Table~\ref{dialogRE-result} compared with other baselines.

Among the models using BERT, TUCORE-GCN$_{BERT}$ outperforms all baselines by 5.3 $\sim$ 7.6 $F1$ scores and 2.9 $\sim$ 7.1 $F1_c$ scores on the test set. GDPNet, the state-of-the-art model, achieved high-performance improvement at $F1_c$, but TUCORE-GCN showed high-performance improvement at both $F1$ and $F1_c$. Among the models using RoBERTa, TUCORE-GCN$_{RoBERTa}$ yields a great improvement of $F1$/$F1_c$ on the test set by 1.8/2.2, in comparison with the strong baseline RoBERTa$_s$. Our model can use BERT (or its variants) as an encoder, and in the experiment, we used both the BERT-base model and also the RoBERTa-large model. TUCORE-GCN show outstanding performance even when the BERT-base was used as the encoder. RoBERTa-large was also used, and it achieves state-of-the-art performance on DialogRE dataset with $F1$ score 73.1 and $F1_c$ score 65.9. It suggests that TUCORE-GCN is very effective in this dialog-based RE task.

\subsubsection{Analysis on Inverse Relations}
\label{ssec:Analysis}
We analyze asymmetric inverse relations and symmetric inverse relations performance on the dialogue-based RE task. We divide the DialogRE dev set into three groups depending on whether it was asymmetric inverse relation, symmetric inverse relation, or other. Then, we report the $F1$ score for each group in Appendix. In the dialogue-based RE task, when asymmetric inverse relations are predicted to exists, BERT makes more mistakes compared to symmetric inverse relations \citep{yu-etal-2020-dialogue}. Since BERT learns the tokenized representation of the input sequence through a Self-Attention mechanism, whether the arguments in the input sequence are a subject or an object is not learned in detail. As a result, the performance of asymmetric inverse relations that indicate different relations when subject and object are changed is significantly lower than in symmetric inverse relations that indicate the same relations even when subject and object are changed. 
However, TUCORE-GCN creates nodes for arguments separately, learns features of these nodes, and classifies relations.
Thus, these issues with BERT can be improved. TUCORE-GCN$_{BERT}$ has improved performance in all groups compared to BERT and BERT$_s$, especially for asymmetric inverse relations.

\subsection{Emotion Recognition in Conversations}
\subsubsection{Dataset}
We evaluate our model on three ERC benchmark datasets. We follow the standard split of the datasets and classify the emotion label of each utterance in the ERC benchmark datasets as the relation between the speaker and the utterance in the dialogue as in Table~\ref{erd2dialogRE-table}. 

\textbf{MELD \citep{poria-etal-2019-meld}\footnote{https://affective-meld.github.io}} is a multimodal dataset collected from the TV show, Friends. We only used textual modality in this dataset. It has seven emotion labels, 2,458 dialogues, and 12,708 utterances. Each utterance is annotated with one of the seven emotion labels. 

\textbf{EmoryNLP \citep{DBLP:conf/aaai/ZahiriC18}\footnote{https://github.com/emorynlp/emotion-detection}} is also collected from the TV show, Friends. It has seven emotion labels, 897 dialogues, and 12,606 utterances. Each utterance is annotated with one of the seven emotion labels. 

\textbf{DailyDialog \citep{li-etal-2017-dailydialog}\footnote{http://yanran.li/dailydialog.html}} reflects our daily communication way and covers various topics about our daily life. It has seven emotion labels, 13,118 dialogues, and 102,979 utterances. Each utterance is annotated with one of the seven emotion labels. Since it does not have speaker information, we consider the utterance turns as two anonymized speaker turns by default.

\begin{table*}
\centering
{\small
\begin{tabular}{c|ccc}
\Xhline{3\arrayrulewidth}
\textbf{Method} & \textbf{MELD} & \textbf{EmoryNLP} & \textbf{DailyDialog}\\
\hline
CNN & 55.86 & 32.59 & 49.34 \\ 
CNN+cLSTM & 56.87 & 32.89 & 50.24 \\ 
DialogueRNN & 57.03 & 31.70 & 50.65 \\ 
DialogueGCN & 58.10 & - & - \\ 
RGAT & 60.91 & 34.42 & 54.31 \\
RoBERTa & 62.02 & 37.29 & 55.16 \\
RoBERTa$_r$ & 64.19 & 38.03 & 61.65 \\
COSMIC & 65.21 & 38.11 & 58.48 \\
CESTa & 58.36 & - & \textbf{63.12} \\
\hline
TUCORE-GCN$_{BERT}$ & 62.47 & 36.01 & 58.34 \\
TUCORE-GCN$_{RoBERTa}$ & \textbf{65.36} & \textbf{39.24} & 61.91 \\
\Xhline{3\arrayrulewidth}
\end{tabular}
}
\caption{\label{ERC-result} Overall performance on three ERC datasets. “-” signifies that no results were reported for the given dataset. Performance scores of TUCORE-GCN$_{BERT}$ on MELD, EmoryNLP, and DailyDialog have standard deviations of 0.4, 0.7, and 0.4, respectively, and performance scores of TUCORE-GCN$_{RoBERTa}$ have standard deviations of 0.4, 0.6, and 0.8, respectively.}
\end{table*}

\subsubsection{Metrics}
For DailyDialog, we calculate micro-$F1$ except for the neutral class, because of its extremely high majority. For MELD and EmoryNLP, we calculate weighted-$F1$.

\subsubsection{Baselines and State-of-the-Art}
For a comprehensive performance evaluation, we compared our model with the models using the following baseline and state-of-the-art methods: 

\textbf{Previous methods:} CNN \citep{kim-2014-convolutional}, CNN+cLSTM \citep{poria-etal-2017-context}, DialogueRNN \citep{MajumderPHMGC19}, DialogueGCN \citep{ghosal-etal-2019-dialoguegcn}, and RoBERTa \citep{liu2019roberta}.

\textbf{RGAT \citep{ishiwatari-etal-2020-relation}:} A model that is provided with some information reflecting relation types through relational position encodings that can capture speaker dependency and sequential information. 

\textbf{RoBERTa$_r$:} The RoBERTa baseline for ERC as our proposed approach, initialized with pre-trained parameters of RoBERTa-large \citep{liu2019roberta}. We set the input sequence of RoBERTa to \texttt{[CLS]d[SEP]a$_1$[SEP]a$_2$[SEP]} and feed them into RoBERTa for classification.

\textbf{COSMIC \citep{ghosal-etal-2020-cosmic}:} A state-of-the-art model for MELD and EmoryNLP. It uses RoBERTa-large as an encoder. It is a framework that models various aspects of commonsense knowledge by considering mental states, events, actions, and cause-effect relations for emotional recognition in conversation.

\textbf{CESTa \citep{wang-etal-2020-contextualized}:} A state-of-the-art model for DailyDialog that uses Conditional Random Field layer. The layer is used for sequential tagging, and it has an advantage in learning when there is an emotional consistency in conversation.

\subsubsection{Results}
We show the performance of TUCORE-GCN on the ERC datasets in Table~\ref{ERC-result}, in comparison with other baselines. 
In addition, the performance of TUCORE-GCN$_{BERT}$ ($F1$($\sigma$)) on the development sets of MELD, EmoryNLP, and DailyDialog is 59.75 (0.5), 37.95 (0.8), 60.25 (0.4), respectively, and the performance of TUCORE-GCN$_{RoBERTa}$ is 65.94 (0.5), 40.17 (0.6), 62.83 (0.5), respectively. We have quoted the results for the baselines and state-of-the-art results reported in \citep{ishiwatari-etal-2020-relation, ghosal-etal-2020-cosmic, wang-etal-2020-contextualized}, except for the results of RoBERTa$_r$.

The only difference between RoBERTa and RoBERTa$_r$ is the form of the input sequence, but RoBERTa$_r$ is better at solving ERC task. Accordingly, we claim that treating the ERC as a dialogue-based RE is useful in practice. 
TUCORE-GCN$_{RoBERTa}$ outperforms COSMIC, the previous state-of-the-art model for MELD and EmoryNLP, by 0.15, 1.13, and 3.43 on test sets of MELD, EmoryNLP, and DailyDialog respectively. It shows state-of-art performance on both MELD and EmoryNLP.
On the other hand, TUCORE-GCN$_{RoBERTa}$ shows slightly lower performance than CESTa on DailyDialog dataset. When utterances in a conversation are adjacent to one another, they tend to show similar emotions. This is called emotional consistency, and CRF layer of CESTa fits well with this tendency. Therefore, it has better performance on DailyDialog, which shows emotional consistency well. However, it shows very poor performance on MELD, where emotional consistency does not appear much \citep{wang-etal-2020-contextualized}.
Considering these observations, our model generally shows outstanding performance on MELD, EmoryNLP, and also DailyDialog. It suggests that TUCORE-GCN is effective in ERC as well as dialogue-based RE.

\begin{table*}
\centering
{\small
\begin{tabular}{l|cc|c|c}
\Xhline{3\arrayrulewidth}
\multirow{2}{*}{\textbf{Method}} & \multicolumn{2}{c|}{\textbf{DialogRE}} & \textbf{MELD} &\textbf{EmoryNLP} \\
& F1 ($\sigma$) & F1$_c$ ($\sigma$) & F1 ($\sigma$) & F1 ($\sigma$) \\
\hline
TUCORE-GCN$_{RoBERTa}$ & 73.1 (0.4) & 65.9 (0.6) & 65.36 (0.4) & 39.24 (0.6) \\
w/o speaker embedding & 72.7 (1.0) & 65.8 (0.6) & 64.64 (0.9) & 37.59 (0.5) \\
w/o turn attention & 72.0 (0.6) & 65.3 (0.3) & 64.59 (0.4) & 37.07 (1.1) \\
w/o turn-level BiLSTM & 72.5 (0.4) & 65.7 (0.3) & 65.02 (0.4) & 38.35 (0.5) \\
\Xhline{3\arrayrulewidth}
\end{tabular}
}
\caption{\label{ablation} Results of ablation study on DialogRE, MELD, and EmoryNLP.}
\end{table*}

\subsection{Ablation Study}
\label{ssec:ablation study}
We conduct ablation studies to evaluate the effectiveness of different modules and mechanisms in TUCORE-GCN. The results are shown in Table~\ref{ablation}.

First, we removed the speaker embedding in the encoder module. To be specific, the encoder and input format of TUCORE-GCN$_{RoBERTa}$ are the same as RoBERTa$_s$. Without speaker embedding, the performance of TUCORE-GCN$_{RoBERTa}$ drops by 0.4 $F1$ score and 0.1 $F1_c$ score on the DialogRE test set and 0.72 and 1.65 $F1$ scores on the MELD and EmoryNLP test set, respectively. This drop shows that when encoding a dialogue, a better representation can be obtained through speaker change information.

Next, we removed the turn attention module. To be specific, the output of the encoding module is delivered to the dialogue graph with sequential nodes module. Without turn attention, the performance of TUCORE-GCN$_{RoBERTa}$ sharply drops by 1.1 $F1$ score and 0.6 $F1_c$ score on DialogRE test set and 0.77 and 2.17 $F1$ scores on the MELD and EmoryNLP test set, respectively. This drop shows that the turn attention module helps capture the representation of the turns and, therefore, improves dialogue-based RE and ERC performance.

Finally, we removed the turn-level BiLSTM for turn nodes in the dialogue graph with sequential nodes module. To be specific, in the module, we apply GCN without injecting sequential information of the turn nodes. Without turn-level BiLSTM, the performance of TUCORE-GCN$_{RoBERTa}$ drops by 0.6 $F1$ score and 0.2 $F1_c$ score on DialogRE test set and 0.34 and 0.89 $F1$ scores on MELD and EmoryNLP test set, respectively. This means that reflecting the characteristics of the sequential nodes when learning the graph helps to learn the features of each node and, therefore, improves dialogue-based RE and ERC performance.

\section{Conclusion and Future Work}
In this paper, we propose TUCORE-GCN for dialogue-based RE. TUCORE-GCN is designed according to the way people understand dialogues in practice to better cope with dialogue-based RE. In addition, we propose a way to treat the ERC task as dialogue-based RE and showed its effectiveness through experiments. Experimental results on a dialogue-based RE dataset and three ERC task datasets demonstrate that TUCORE-GCN model significantly outperforms existing models and yields the new state-of-the-art results on both tasks.

Since TUCORE-GCN is modeled for the dialogue text type, we expect it to perform well in dialogue-based natural language understanding tasks. In future work, we are going to explore the effectiveness of it on other dialogue-based tasks.

\section*{Acknowledgements}
We would like to thank Jinwoo Jeong, Tae Hee Jo, Sangwoo Seo, and the anonymous reviewers for their thoughtful and constructive comments. 
This work was supported by Institute of Information \& communications Technology Planning \& Evaluation(IITP) grant funded by the Korea government(MSIT) (No. 2020-0-01373, Artificial Intelligence Graduate School Program (Hanyang University)) and the National Research Foundation of Korea(NRF) grant funded by the Korea government(*MSIT) (No.2018R1A5A7059549). *Ministry of Science and ICT.

\bibliography{anthology,custom}

\begin{thebibliography}{39}
\expandafter\ifx\csname natexlab\endcsname\relax\def\natexlab#1{#1}\fi

\bibitem[{Biber(1988)}]{Biber_Variation88}
Douglas Biber. 1988.
\newblock \emph{Variation across Speech and Writing}.
\newblock Cambridge University Press, Cambridge.

\bibitem[{Christopoulou et~al.(2019)Christopoulou, Miwa, and
  Ananiadou}]{christopoulou-etal-2019-connecting}
Fenia Christopoulou, Makoto Miwa, and Sophia Ananiadou. 2019.
\newblock \href {https://doi.org/10.18653/v1/D19-1498} {Connecting the dots:
  Document-level neural relation extraction with edge-oriented graphs}.
\newblock In \emph{Proceedings of the 2019 Conference on Empirical Methods in
  Natural Language Processing and the 9th International Joint Conference on
  Natural Language Processing (EMNLP-IJCNLP)}, pages 4925--4936, Hong Kong,
  China. Association for Computational Linguistics.

\bibitem[{Conneau and Lample(2019)}]{NEURIPS2019_c04c19c2}
Alexis Conneau and Guillaume Lample. 2019.
\newblock \href
  {https://proceedings.neurips.cc/paper/2019/hash/c04c19c2c2474dbf5f7ac4372c5b9af1-Abstract.html}
  {Cross-lingual language model pretraining}.
\newblock In \emph{Advances in Neural Information Processing Systems 32: Annual
  Conference on Neural Information Processing Systems 2019, NeurIPS 2019,
  December 8-14, 2019, Vancouver, BC, Canada}, pages 7057--7067.

\bibitem[{Devlin et~al.(2019)Devlin, Chang, Lee, and
  Toutanova}]{devlin-etal-2019-bert}
Jacob Devlin, Ming-Wei Chang, Kenton Lee, and Kristina Toutanova. 2019.
\newblock \href {https://doi.org/10.18653/v1/N19-1423} {{BERT}: Pre-training of
  deep bidirectional transformers for language understanding}.
\newblock In \emph{Proceedings of the 2019 Conference of the North {A}merican
  Chapter of the Association for Computational Linguistics: Human Language
  Technologies, Volume 1 (Long and Short Papers)}, pages 4171--4186,
  Minneapolis, Minnesota. Association for Computational Linguistics.

\bibitem[{Ghosal et~al.(2020)Ghosal, Majumder, Gelbukh, Mihalcea, and
  Poria}]{ghosal-etal-2020-cosmic}
Deepanway Ghosal, Navonil Majumder, Alexander Gelbukh, Rada Mihalcea, and
  Soujanya Poria. 2020.
\newblock \href {https://doi.org/10.18653/v1/2020.findings-emnlp.224}
  {{COSMIC}: {CO}mmon{S}ense knowledge for e{M}otion identification in
  conversations}.
\newblock In \emph{Findings of the Association for Computational Linguistics:
  EMNLP 2020}, pages 2470--2481, Online. Association for Computational
  Linguistics.

\bibitem[{Ghosal et~al.(2019)Ghosal, Majumder, Poria, Chhaya, and
  Gelbukh}]{ghosal-etal-2019-dialoguegcn}
Deepanway Ghosal, Navonil Majumder, Soujanya Poria, Niyati Chhaya, and
  Alexander Gelbukh. 2019.
\newblock \href {https://doi.org/10.18653/v1/D19-1015} {{D}ialogue{GCN}: A
  graph convolutional neural network for emotion recognition in conversation}.
\newblock In \emph{Proceedings of the 2019 Conference on Empirical Methods in
  Natural Language Processing and the 9th International Joint Conference on
  Natural Language Processing (EMNLP-IJCNLP)}, pages 154--164, Hong Kong,
  China. Association for Computational Linguistics.

\bibitem[{Gu et~al.(2020)Gu, Li, Liu, Ling, Su, Wei, and Zhu}]{gu2020speaker}
Jia{-}Chen Gu, Tianda Li, Quan Liu, Zhen{-}Hua Ling, Zhiming Su, Si~Wei, and
  Xiaodan Zhu. 2020.
\newblock \href {https://doi.org/10.1145/3340531.3412330} {Speaker-aware {BERT}
  for multi-turn response selection in retrieval-based chatbots}.
\newblock In \emph{{CIKM} '20: The 29th {ACM} International Conference on
  Information and Knowledge Management, Virtual Event, Ireland, October 19-23,
  2020}, pages 2041--2044. {ACM}.

\bibitem[{Guo et~al.(2019)Guo, Cao, Zhang, Shu, and Liu}]{guo2019dean}
Chuan Guo, Juan Cao, Xueyao Zhang, Kai Shu, and Huan Liu. 2019.
\newblock \href {https://arxiv.org/abs/1903.01728} {Dean: Learning dual emotion
  for fake news detection on social media}.
\newblock \emph{ArXiv preprint}, abs/1903.01728.

\bibitem[{Ishiwatari et~al.(2020)Ishiwatari, Yasuda, Miyazaki, and
  Goto}]{ishiwatari-etal-2020-relation}
Taichi Ishiwatari, Yuki Yasuda, Taro Miyazaki, and Jun Goto. 2020.
\newblock \href {https://doi.org/10.18653/v1/2020.emnlp-main.597}
  {Relation-aware graph attention networks with relational position encodings
  for emotion recognition in conversations}.
\newblock In \emph{Proceedings of the 2020 Conference on Empirical Methods in
  Natural Language Processing (EMNLP)}, pages 7360--7370, Online. Association
  for Computational Linguistics.

\bibitem[{Ji et~al.(2010)Ji, Grishman, Dang, Griffitt, and
  Ellis}]{Ji10overviewof}
Heng Ji, Ralph Grishman, Hoa~Trang Dang, Kira Griffitt, and Joe Ellis. 2010.
\newblock Overview of the tac 2010 knowledge base population track.
\newblock In \emph{In Third Text Analysis Conference}.

\bibitem[{Kim(2014)}]{kim-2014-convolutional}
Yoon Kim. 2014.
\newblock \href {https://doi.org/10.3115/v1/D14-1181} {Convolutional neural
  networks for sentence classification}.
\newblock In \emph{Proceedings of the 2014 Conference on Empirical Methods in
  Natural Language Processing ({EMNLP})}, pages 1746--1751, Doha, Qatar.
  Association for Computational Linguistics.

\bibitem[{Kingma and Ba(2015)}]{DBLP:journals/corr/KingmaB14}
Diederik~P. Kingma and Jimmy Ba. 2015.
\newblock \href {http://arxiv.org/abs/1412.6980} {Adam: {A} method for
  stochastic optimization}.
\newblock In \emph{3rd International Conference on Learning Representations,
  {ICLR} 2015, San Diego, CA, USA, May 7-9, 2015, Conference Track
  Proceedings}.

\bibitem[{Kipf and Welling(2017)}]{Kipf:2016tc}
Thomas~N. Kipf and Max Welling. 2017.
\newblock \href {https://openreview.net/forum?id=SJU4ayYgl} {Semi-supervised
  classification with graph convolutional networks}.
\newblock In \emph{5th International Conference on Learning Representations,
  {ICLR} 2017, Toulon, France, April 24-26, 2017, Conference Track
  Proceedings}. OpenReview.net.

\bibitem[{Lan et~al.(2020)Lan, Chen, Goodman, Gimpel, Sharma, and
  Soricut}]{Lan2020ALBERT:}
Zhenzhong Lan, Mingda Chen, Sebastian Goodman, Kevin Gimpel, Piyush Sharma, and
  Radu Soricut. 2020.
\newblock \href {https://openreview.net/forum?id=H1eA7AEtvS} {{ALBERT:} {A}
  lite {BERT} for self-supervised learning of language representations}.
\newblock In \emph{8th International Conference on Learning Representations,
  {ICLR} 2020, Addis Ababa, Ethiopia, April 26-30, 2020}. OpenReview.net.

\bibitem[{Lee and Hong(2016)}]{LEE2016360}
Jieun Lee and Ilyoo~B. Hong. 2016.
\newblock \href
  {https://doi.org/https://doi.org/10.1016/j.ijinfomgt.2016.01.001} {Predicting
  positive user responses to social media advertising: The roles of emotional
  appeal, informativeness, and creativity}.
\newblock \emph{International Journal of Information Management},
  36(3):360--373.

\bibitem[{Li et~al.(2017)Li, Su, Shen, Li, Cao, and
  Niu}]{li-etal-2017-dailydialog}
Yanran Li, Hui Su, Xiaoyu Shen, Wenjie Li, Ziqiang Cao, and Shuzi Niu. 2017.
\newblock \href {https://aclanthology.org/I17-1099} {{D}aily{D}ialog: A
  manually labelled multi-turn dialogue dataset}.
\newblock In \emph{Proceedings of the Eighth International Joint Conference on
  Natural Language Processing (Volume 1: Long Papers)}, pages 986--995, Taipei,
  Taiwan. Asian Federation of Natural Language Processing.

\bibitem[{Liu et~al.(2019)Liu, Ott, Goyal, Du, Joshi, Chen, Levy, Lewis,
  Zettlemoyer, and Stoyanov}]{liu2019roberta}
Yinhan Liu, Myle Ott, Naman Goyal, Jingfei Du, Mandar Joshi, Danqi Chen, Omer
  Levy, Mike Lewis, Luke Zettlemoyer, and Veselin Stoyanov. 2019.
\newblock \href {https://arxiv.org/abs/1907.11692} {Roberta: A robustly
  optimized bert pretraining approach}.
\newblock \emph{ArXiv preprint}, abs/1907.11692.

\bibitem[{Majumder et~al.(2019)Majumder, Poria, Hazarika, Mihalcea, Gelbukh,
  and Cambria}]{MajumderPHMGC19}
Navonil Majumder, Soujanya Poria, Devamanyu Hazarika, Rada Mihalcea,
  Alexander~F. Gelbukh, and Erik Cambria. 2019.
\newblock \href {https://doi.org/10.1609/aaai.v33i01.33016818} {Dialoguernn: An
  attentive {RNN} for emotion detection in conversations}.
\newblock In \emph{The Thirty-Third {AAAI} Conference on Artificial
  Intelligence, {AAAI} 2019, The Thirty-First Innovative Applications of
  Artificial Intelligence Conference, {IAAI} 2019, The Ninth {AAAI} Symposium
  on Educational Advances in Artificial Intelligence, {EAAI} 2019, Honolulu,
  Hawaii, USA, January 27 - February 1, 2019}, pages 6818--6825. {AAAI} Press.

\bibitem[{Nan et~al.(2020)Nan, Guo, Sekulic, and Lu}]{nan-etal-2020-reasoning}
Guoshun Nan, Zhijiang Guo, Ivan Sekulic, and Wei Lu. 2020.
\newblock \href {https://doi.org/10.18653/v1/2020.acl-main.141} {Reasoning with
  latent structure refinement for document-level relation extraction}.
\newblock In \emph{Proceedings of the 58th Annual Meeting of the Association
  for Computational Linguistics}, pages 1546--1557, Online. Association for
  Computational Linguistics.

\bibitem[{Poria et~al.(2017)Poria, Cambria, Hazarika, Majumder, Zadeh, and
  Morency}]{poria-etal-2017-context}
Soujanya Poria, Erik Cambria, Devamanyu Hazarika, Navonil Majumder, Amir Zadeh,
  and Louis-Philippe Morency. 2017.
\newblock \href {https://doi.org/10.18653/v1/P17-1081} {Context-dependent
  sentiment analysis in user-generated videos}.
\newblock In \emph{Proceedings of the 55th Annual Meeting of the Association
  for Computational Linguistics (Volume 1: Long Papers)}, pages 873--883,
  Vancouver, Canada. Association for Computational Linguistics.

\bibitem[{Poria et~al.(2019)Poria, Hazarika, Majumder, Naik, Cambria, and
  Mihalcea}]{poria-etal-2019-meld}
Soujanya Poria, Devamanyu Hazarika, Navonil Majumder, Gautam Naik, Erik
  Cambria, and Rada Mihalcea. 2019.
\newblock \href {https://doi.org/10.18653/v1/P19-1050} {{MELD}: A multimodal
  multi-party dataset for emotion recognition in conversations}.
\newblock In \emph{Proceedings of the 57th Annual Meeting of the Association
  for Computational Linguistics}, pages 527--536, Florence, Italy. Association
  for Computational Linguistics.

\bibitem[{{Schuster} and {Paliwal}(1997)}]{650093}
M.~{Schuster} and K.~K. {Paliwal}. 1997.
\newblock \href {https://doi.org/10.1109/78.650093} {Bidirectional recurrent
  neural networks}.
\newblock \emph{IEEE Transactions on Signal Processing}, 45(11):2673--2681.

\bibitem[{Surdeanu(2013)}]{Surdeanu2013OverviewOT}
M.~Surdeanu. 2013.
\newblock Overview of the tac2013 knowledge base population evaluation: English
  slot filling and temporal slot filling.
\newblock \emph{Theory and Applications of Categories}.

\bibitem[{Swampillai and Stevenson(2010)}]{swampillai-stevenson-2010-inter}
Kumutha Swampillai and Mark Stevenson. 2010.
\newblock \href
  {http://www.lrec-conf.org/proceedings/lrec2010/pdf/905_Paper.pdf}
  {Inter-sentential relations in information extraction corpora}.
\newblock In \emph{Proceedings of the Seventh International Conference on
  Language Resources and Evaluation ({LREC}'10)}, Valletta, Malta. European
  Language Resources Association (ELRA).

\bibitem[{Tang et~al.(2020)Tang, Cao, Zhang, Cao, Fang, Wang, and
  Yin}]{DBLP:conf/pakdd/TangC0CFWY20}
Hengzhu Tang, Yanan Cao, Zhenyu Zhang, Jiangxia Cao, Fang Fang, Shi Wang, and
  Pengfei Yin. 2020.
\newblock \href {https://doi.org/10.1007/978-3-030-47426-3\_16} {{HIN:}
  hierarchical inference network for document-level relation extraction}.
\newblock In \emph{Advances in Knowledge Discovery and Data Mining - 24th
  Pacific-Asia Conference, {PAKDD} 2020, Singapore, May 11-14, 2020,
  Proceedings, Part {I}}, volume 12084 of \emph{Lecture Notes in Computer
  Science}, pages 197--209. Springer.

\bibitem[{Vaswani et~al.(2017)Vaswani, Shazeer, Parmar, Uszkoreit, Jones,
  Gomez, Kaiser, and Polosukhin}]{NIPS2017_3f5ee243}
Ashish Vaswani, Noam Shazeer, Niki Parmar, Jakob Uszkoreit, Llion Jones,
  Aidan~N. Gomez, Lukasz Kaiser, and Illia Polosukhin. 2017.
\newblock \href
  {https://proceedings.neurips.cc/paper/2017/hash/3f5ee243547dee91fbd053c1c4a845aa-Abstract.html}
  {Attention is all you need}.
\newblock In \emph{Advances in Neural Information Processing Systems 30: Annual
  Conference on Neural Information Processing Systems 2017, December 4-9, 2017,
  Long Beach, CA, {USA}}, pages 5998--6008.

\bibitem[{Wang and Liu(2011)}]{wang-liu-2011-pilot}
Dong Wang and Yang Liu. 2011.
\newblock \href {https://aclanthology.org/P11-1034} {A pilot study of opinion
  summarization in conversations}.
\newblock In \emph{Proceedings of the 49th Annual Meeting of the Association
  for Computational Linguistics: Human Language Technologies}, pages 331--339,
  Portland, Oregon, USA. Association for Computational Linguistics.

\bibitem[{Wang et~al.(2019)Wang, Focke, Sylvester, Mishra, and
  Wang}]{DBLP:journals/corr/abs-1909-11898}
Hong Wang, Christfried Focke, Rob Sylvester, Nilesh Mishra, and William~Yang
  Wang. 2019.
\newblock \href {http://arxiv.org/abs/1909.11898} {Fine-tune bert for docred
  with two-step process}.
\newblock \emph{CoRR}, abs/1909.11898.

\bibitem[{Wang et~al.(2016)Wang, Cao, de~Melo, and
  Liu}]{wang-etal-2016-relation}
Linlin Wang, Zhu Cao, Gerard de~Melo, and Zhiyuan Liu. 2016.
\newblock \href {https://doi.org/10.18653/v1/P16-1123} {Relation classification
  via multi-level attention {CNN}s}.
\newblock In \emph{Proceedings of the 54th Annual Meeting of the Association
  for Computational Linguistics (Volume 1: Long Papers)}, pages 1298--1307,
  Berlin, Germany. Association for Computational Linguistics.

\bibitem[{Wang et~al.(2020)Wang, Zhang, Ma, Wang, and
  Xiao}]{wang-etal-2020-contextualized}
Yan Wang, Jiayu Zhang, Jun Ma, Shaojun Wang, and Jing Xiao. 2020.
\newblock \href {https://aclanthology.org/2020.sigdial-1.23} {Contextualized
  emotion recognition in conversation as sequence tagging}.
\newblock In \emph{Proceedings of the 21th Annual Meeting of the Special
  Interest Group on Discourse and Dialogue}, pages 186--195, 1st virtual
  meeting. Association for Computational Linguistics.

\bibitem[{Xue et~al.(2021)Xue, Sun, Zhang, and Chng}]{DBLP:conf/aaai/XueSZC21}
Fuzhao Xue, Aixin Sun, Hao Zhang, and Eng~Siong Chng. 2021.
\newblock \href {https://ojs.aaai.org/index.php/AAAI/article/view/17670}
  {Gdpnet: Refining latent multi-view graph for relation extraction}.
\newblock In \emph{Thirty-Fifth {AAAI} Conference on Artificial Intelligence,
  {AAAI} 2021, Thirty-Third Conference on Innovative Applications of Artificial
  Intelligence, {IAAI} 2021, The Eleventh Symposium on Educational Advances in
  Artificial Intelligence, {EAAI} 2021, Virtual Event, February 2-9, 2021},
  pages 14194--14202. {AAAI} Press.

\bibitem[{Yao et~al.(2019)Yao, Ye, Li, Han, Lin, Liu, Liu, Huang, Zhou, and
  Sun}]{yao-etal-2019-docred}
Yuan Yao, Deming Ye, Peng Li, Xu~Han, Yankai Lin, Zhenghao Liu, Zhiyuan Liu,
  Lixin Huang, Jie Zhou, and Maosong Sun. 2019.
\newblock \href {https://doi.org/10.18653/v1/P19-1074} {{D}oc{RED}: A
  large-scale document-level relation extraction dataset}.
\newblock In \emph{Proceedings of the 57th Annual Meeting of the Association
  for Computational Linguistics}, pages 764--777, Florence, Italy. Association
  for Computational Linguistics.

\bibitem[{Ye et~al.(2020)Ye, Lin, Du, Liu, Li, Sun, and
  Liu}]{Ye2020CoreferentialRL}
Deming Ye, Yankai Lin, Jiaju Du, Zhenghao Liu, Peng Li, Maosong Sun, and
  Zhiyuan Liu. 2020.
\newblock \href {https://doi.org/10.18653/v1/2020.emnlp-main.582}
  {{C}oreferential {R}easoning {L}earning for {L}anguage {R}epresentation}.
\newblock In \emph{Proceedings of the 2020 Conference on Empirical Methods in
  Natural Language Processing (EMNLP)}, pages 7170--7186, Online. Association
  for Computational Linguistics.

\bibitem[{Yu et~al.(2020)Yu, Sun, Cardie, and Yu}]{yu-etal-2020-dialogue}
Dian Yu, Kai Sun, Claire Cardie, and Dong Yu. 2020.
\newblock \href {https://doi.org/10.18653/v1/2020.acl-main.444} {Dialogue-based
  relation extraction}.
\newblock In \emph{Proceedings of the 58th Annual Meeting of the Association
  for Computational Linguistics}, pages 4927--4940, Online. Association for
  Computational Linguistics.

\bibitem[{Zahiri and Choi(2018)}]{DBLP:conf/aaai/ZahiriC18}
Sayyed~M. Zahiri and Jinho~D. Choi. 2018.
\newblock \href {https://aaai.org/ocs/index.php/WS/AAAIW18/paper/view/16434}
  {Emotion detection on {TV} show transcripts with sequence-based convolutional
  neural networks}.
\newblock In \emph{The Workshops of the The Thirty-Second {AAAI} Conference on
  Artificial Intelligence, New Orleans, Louisiana, USA, February 2-7, 2018},
  volume {WS-18} of \emph{{AAAI} Workshops}, pages 44--52. {AAAI} Press.

\bibitem[{Zeng et~al.(2014)Zeng, Liu, Lai, Zhou, and
  Zhao}]{zeng-etal-2014-relation}
Daojian Zeng, Kang Liu, Siwei Lai, Guangyou Zhou, and Jun Zhao. 2014.
\newblock \href {https://aclanthology.org/C14-1220} {Relation classification
  via convolutional deep neural network}.
\newblock In \emph{Proceedings of {COLING} 2014, the 25th International
  Conference on Computational Linguistics: Technical Papers}, pages 2335--2344,
  Dublin, Ireland. Dublin City University and Association for Computational
  Linguistics.

\bibitem[{Zeng et~al.(2020)Zeng, Xu, Chang, and Li}]{zeng-etal-2020-double}
Shuang Zeng, Runxin Xu, Baobao Chang, and Lei Li. 2020.
\newblock \href {https://doi.org/10.18653/v1/2020.emnlp-main.127} {Double graph
  based reasoning for document-level relation extraction}.
\newblock In \emph{Proceedings of the 2020 Conference on Empirical Methods in
  Natural Language Processing (EMNLP)}, pages 1630--1640, Online. Association
  for Computational Linguistics.

\bibitem[{Zhang et~al.(2017)Zhang, Zhong, Chen, Angeli, and
  Manning}]{zhang-etal-2017-position}
Yuhao Zhang, Victor Zhong, Danqi Chen, Gabor Angeli, and Christopher~D.
  Manning. 2017.
\newblock \href {https://doi.org/10.18653/v1/D17-1004} {Position-aware
  attention and supervised data improve slot filling}.
\newblock In \emph{Proceedings of the 2017 Conference on Empirical Methods in
  Natural Language Processing}, pages 35--45, Copenhagen, Denmark. Association
  for Computational Linguistics.

\bibitem[{Zhu et~al.(2019)Zhu, Lin, Liu, Fu, Chua, and
  Sun}]{zhu-etal-2019-graph}
Hao Zhu, Yankai Lin, Zhiyuan Liu, Jie Fu, Tat-Seng Chua, and Maosong Sun. 2019.
\newblock \href {https://doi.org/10.18653/v1/P19-1128} {Graph neural networks
  with generated parameters for relation extraction}.
\newblock In \emph{Proceedings of the 57th Annual Meeting of the Association
  for Computational Linguistics}, pages 1331--1339, Florence, Italy.
  Association for Computational Linguistics.

\end{thebibliography}
\bibliographystyle{acl_natbib}

\appendix
\section{Appendix}

\subsection{Input Representation}
\begin{figure*}
\begin{center}
\includegraphics[width=\linewidth]{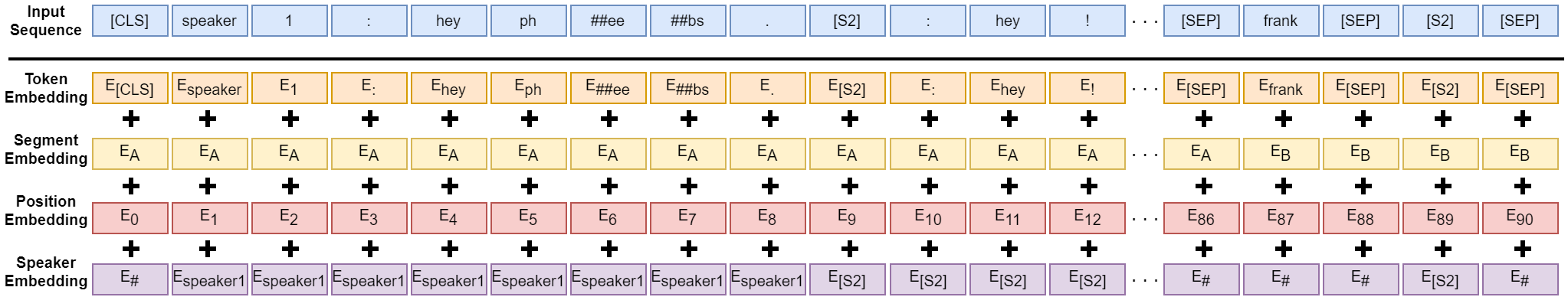}
\end{center}
\caption{The input representation of TUCORE-GCN. The final input embeddings are the sum of the token embeddings, the segment embeddings, the position embeddings and the speaker embeddings.}
\label{fig:speaker embedding}
\end{figure*}

A visual architecture of our input representation is illustrated in Figure~\ref{fig:speaker embedding}.

\subsection{Surrounding Turn Mask}
\begin{figure}
\begin{center}
\includegraphics[width=\linewidth]{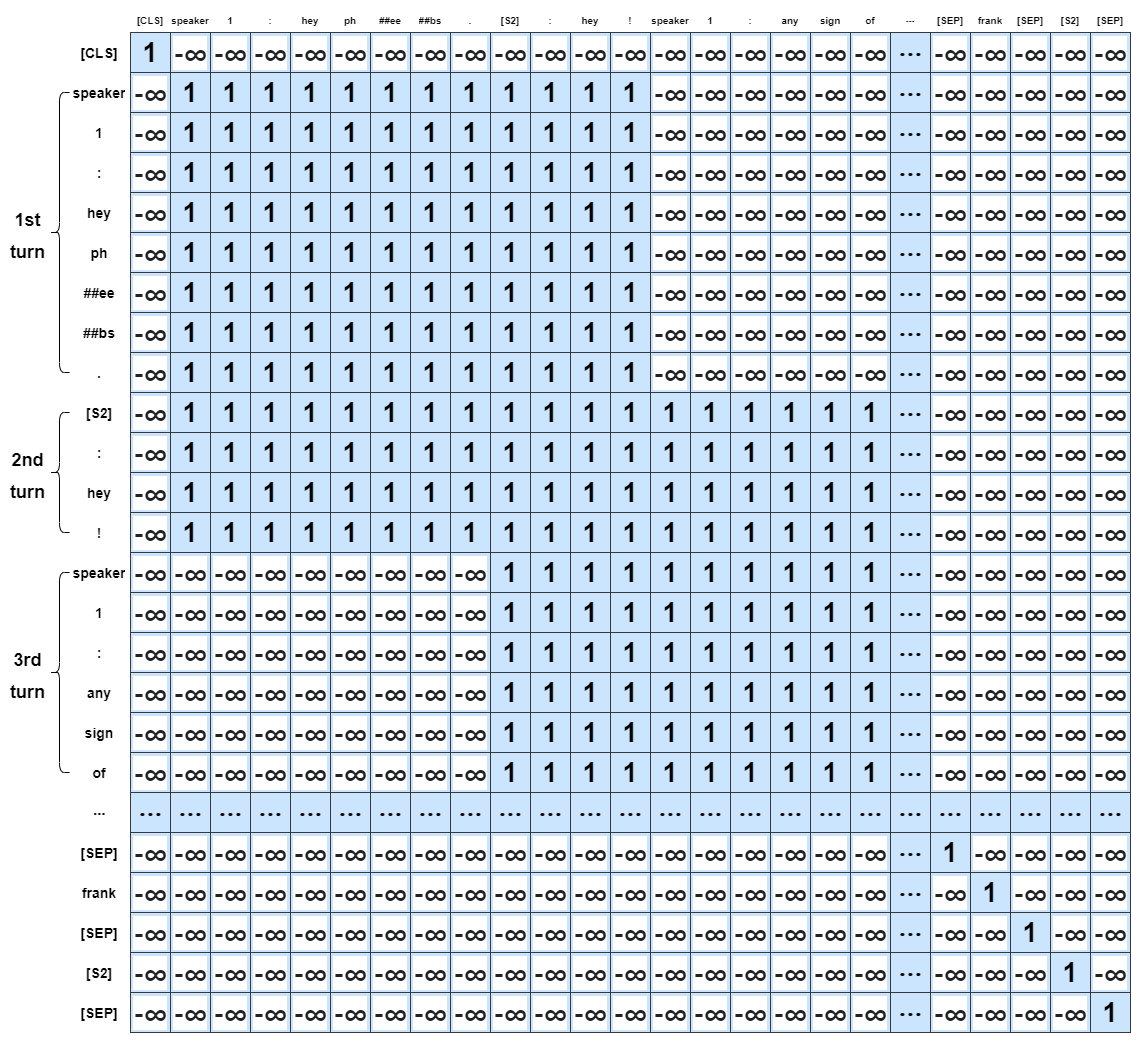}
\end{center}
\caption{When the surround turn window size is 1, it is the surrounding turn mask of TUCORE-GCN. For each token, the surrounding turn and its own turn correspond to 1, and the rest is $-\infty$.}
\label{fig:turn mask}
\end{figure}

A visual architecture of our surrounding turn mask of Turn Attention Module is illustrated in Figure~\ref{fig:turn mask}. 1 was given as an example for surround turn window size c .

\subsection{Experimental Settings}
\subsubsection{Hyperparameter Settings}
We truncate a dialogue when the input sequence length exceeds 512 and use the development set to manually tune the optimal hyperparameters for TUCORE-GCN, based on the $F1$ score. Hyperparameter settings for TUCORE-GCN on a dialogue-based RE dataset are listed in Table~\ref{Setting_dialogRE} and the ones on ERC datasets are listed in Table~\ref{Setting_ERC}. The final values of hyperparameters we adopted are in bold. We do not tune all the hyperparameters.

\subsubsection{Other Settings}
TUCORE-GCN$_{BERT}$ is implemented by using PyTorch 1.6.0 with CUDA 10.1 and TUCORE-GCN$_{RoBERTa}$ is implemented by using PyTorch 1.7.0 with CUDA 11.0. Our implementation of TUCORE-GCN$_{BERT}$ uses the DGL\footnote{\label{note}https://www.dgl.ai} 0.4.3 and our implementation of TUCORE-GCN$_{RoBERTa}$ uses the DGL\textsuperscript{\ref{note}} 0.5.3. We used the official code\footnote{https://github.com/nlpdata/dialogre} of \citep{yu-etal-2020-dialogue} to calculate $F1$ and $F1_c$ scores on DialogRE, and scikit-learn\footnote{https://scikit-learn.org/stable} to calculate $F1$ score on ERC datasets. It takes about 2 hours, 1.25 hours, 1.5 hours, 12 hours to run TUCORE-GCN$_{BERT}$ on DialogRE, MELD, EmoryNLP, and DailyDialog once, respectively. Additionally, it takes about 4 hours, 2.2 hours, 2.3 hours, 20 hours to run TUCORE-GCN$_{RoBERTa}$ on DialogRE, MELD, EmoryNLP, and DailyDialog once, respectively. We conducted all experiments that uses TUCORE-GCN$_{BERT}$ on a Ubuntu server using Intel(R) Core(TM) i9-10900X CPU with 128GB of memory, and used GeForce RTX 2080 Ti GPU with 11GB of memory. We conducted all experiments that uses TUCORE-GCN$_{RoBERTa}$ on a Ubuntu server using Intel(R) Core(TM) i9-10980XE CPU with 128GB of memory, and used GeForce RTX 3090 GPU with 24GB of memory.

\subsection{Experimental results on Inverse Relations}
We show the performance of TUCORE-GCN on asymmetric inverse relation, symmetric inverse relation, and other of DialogRE \citep{yu-etal-2020-dialogue} in Table~\ref{analysis} compared with other baselines. 

Among the models using BERT \citep{devlin-etal-2019-bert}, TUCORE-GCN$_{BERT}$ has significantly reduced difference between the $F1$ scores of asymmetric relation group and the symmetric relation group. The $F1$ score difference between two groups were 5.8, which was the smallest $F1$ score difference compared with the other models that use BERT. In addition, compared to RoBERTa$_s$'s $F1$ score difference between asymmetric relation group and the symmetric relation group, TUCORE-GCN$_{RoBERTa}$'s $F1$ score difference was reduced by 2.9. This suggests that TUCORE-GCN solves the limitations of BERT and its variants' inability to predict the inverse relation well.

\begin{table}
\centering
{\scriptsize
\begin{tabular}{l|c|c}
\Xhline{3\arrayrulewidth}
\multirow{2}{*}{\textbf{Hyperparameter}} & \multicolumn{2}{c}{\textbf{Value}}\\\cline{2-3}
& BERT & RoBERTa\\
\Xhline{3\arrayrulewidth}
Epoch & \textbf{20} & 20, \textbf{30} \\ 
Batch Size & \textbf{12} & \textbf{12} \\
Learning Rate & \textbf{3e-5} & 3e-5, 1e-5, \textbf{5e-6}, 1e-6 \\
Speaker Embedding Size & \textbf{768} & \textbf{768} \\
Layers of Turn Attention & \textbf{1} & \textbf{1} \\
Heads of Turn Attention & \textbf{12} & \textbf{12} \\
Surround Turn Window Size & \textbf{1}, 2, 3 & \textbf{1}, 2 \\
Dropout of Turn Attention & \textbf{0.1} & \textbf{0.1} \\
Layers of LSTM & 1, \textbf{2}, 3 & 1, \textbf{2}, 3 \\
LSTM Hidden Size & \textbf{768} & \textbf{768} \\
Dropout of LSTM & \textbf{0.2}, 0.4, 0.6 & \textbf{0.2}, 0.4 \\
Layers of GCN & \textbf{2} & \textbf{2} \\
GCN Hidden Size & \textbf{768} & \textbf{768}\\
Dropout of GCN & \textbf{0.6} & \textbf{0.6} \\
\hline
\hline
Numbers of Parameters & 156M & 401M \\
Hyperparameter Search Trials & 12 & 12 \\
\Xhline{3\arrayrulewidth}
\end{tabular}
}
\caption{\label{Setting_dialogRE}Settings for TUCORE-GCN$_{BERT}$ and TUCORE-GCN$_{RoBERTa}$ on dialogue-based RE dataset.}
\end{table}

\begin{table}
\centering
{\scriptsize
\begin{tabular}{l|c|c}
\Xhline{3\arrayrulewidth}
\multirow{2}{*}{\textbf{Hyperparameter}} & \multicolumn{2}{c}{\textbf{Value}}\\\cline{2-3}
& BERT & RoBERTa\\
\Xhline{3\arrayrulewidth}
Epoch & \textbf{10} & \textbf{10} \\ 
Batch Size & \textbf{12} & \textbf{12} \\
Learning Rate & \textbf{3e-5} & 3e-5, 1e-5, \textbf{5e-6}, 1e-6 \\
Speaker Embedding Size & \textbf{768} & \textbf{768} \\
Layers of Turn Attention & \textbf{1} & \textbf{1} \\
Heads of Turn Attention & \textbf{12} & \textbf{12} \\
Surround Turn Window Size & \textbf{1} & \textbf{1}, 2 \\
Dropout of Turn Attention & \textbf{0.1} & \textbf{0.1} \\
Layers of LSTM & 1, \textbf{2}, 3 & \textbf{2} \\
LSTM Hidden Size & \textbf{768} & \textbf{768} \\
Dropout of LSTM & \textbf{0.2}, 0.4 & \textbf{0.2} \\
Layers of GCN & \textbf{2} & \textbf{2} \\
GCN Hidden Size & \textbf{768} & \textbf{768} \\
Dropout of GCN & \textbf{0.6} & \textbf{0.6} \\
\hline
\hline
Numbers of Parameters & 156M & 401M \\
Hyperparameter Search Trials & 6 & 6 \\
\Xhline{3\arrayrulewidth}
\end{tabular}
}
\caption{\label{Setting_ERC}Settings for TUCORE-GCN$_{BERT}$ and TUCORE-GCN$_{RoBERTa}$ on ERC datasets.}
\end{table}

\begin{table*}
\centering
{\small
\begin{tabular}{c|ccc}
\Xhline{3\arrayrulewidth}
\textbf{Method} & \textbf{Asymmetric} & \textbf{Symmetric} & \textbf{Other}\\
\hline
BERT \citep{devlin-etal-2019-bert} & 42.5 (3.2) & 60.7 (1.2) & 65.6 (0.8) \\ 
BERT$_s$ \citep{yu-etal-2020-dialogue} & 46.5 (3.3) & 61.5 (0.7) & 69.4 (0.3) \\ 
GDPNet \citep{DBLP:conf/aaai/XueSZC21} & 47.4 (1.9) & 59.8 (2.5) & 68.1 (0.8) \\ 
RoBERTa$_s$ & 57.4 (3.2) & 69.3 (2.1) & 79.6 (1.3) \\ 
\hline
TUCORE-GCN$_{BERT}$ & 57.9 (1.9) & 63.7 (1.9) & 72.2 (0.6) \\
TUCORE-GCN$_{RoBERTa}$ & 62.3 (3.1) & 71.3 (0.8) & 79.9 (0.4) \\
\Xhline{3\arrayrulewidth}
\end{tabular}
}
\caption{\label{analysis} Performance ($F1$ ($\sigma$)) on asymmetric inverse relations group, symmetric inverse relations group, and other relations group of DialogRE \citep{yu-etal-2020-dialogue}. The scores of BERT, BERT$_s$, and GDPNet are based on our re-implementation.}
\end{table*}

\end{document}